\documentclass[runningheads]{llncs}

\usepackage[final,year=2026,ID=6]{eccv}

\usepackage{eccvabbrv}

\usepackage{graphicx}
\usepackage{booktabs}

\usepackage[accsupp]{axessibility}  

\usepackage{hyperref}

\usepackage{orcidlink}

\begin{document}

\title{How Does SAM-3 Adapt? \\Capacity, Update Geometry, and Domain Transfer} 

\titlerunning{How Does SAM-3 Adapt?}

\author{Allan Kazakov \and
Abdurrahman Javat \and
Duygu Cakir}

\authorrunning{A.~Kazakov et al.}

\institute{Bahcesehir University, Istanbul, Turkey\\
\email{\{allan.kazakov, abdurrahman.javat\}@bahcesehir.edu.tr},\\
\email{duygu.cakir@bau.edu.tr}}

\maketitle

\begin{abstract}
Promptable segmentation foundation models are increasingly used in specialized domains, but it remains unclear how SAM-3 should be fine-tuned when adapting it beyond its original training distribution. We study this question by comparing five adaptation families---LoRA, HiRA, KRAdapter, QuanTA, and Full Fine-Tuning---across five segmentation datasets spanning medical imaging, remote sensing, and dense material segmentation. Across these datasets, compact parameter-efficient fine-tuning methods often match or approach Full Fine-Tuning in DSC and IoU, suggesting that strong in-domain adaptation does not always require updating the full model. However, this performance cannot be explained by nominal adapter rank or trainable parameter count alone. To investigate why different methods perform similarly, we analyze their update spectra and their behavior under LoveDA Urban/Rural transfer. Our SVD analysis shows that Full Fine-Tuning updates are spectrally compressible. However, successful PEFT methods do not simply recover the same update directions; instead, they often concentrate changes in different components and subspaces. The domain-transfer study further shows that methods with similar in-domain performance can respond differently under distribution shift. Together, these results suggest that SAM-3 adaptation is often capacity-saturated in practice, while competitive fine-tuning arises through method-specific update paths rather than a single low-rank approximation to Full Fine-Tuning.
\end{abstract}

\section{Introduction}

Promptable segmentation foundation models have made it possible to adapt a single pretrained model to many downstream segmentation tasks with minimal task-specific design. However, as these models become larger and more modular, fine-tuning them becomes a nontrivial design choice. Full fine-tuning offers maximum adaptation capacity, but it is costly and may be unnecessary for many specialized domains. Parameter-efficient fine-tuning (PEFT) methods provide a more practical alternative, yet it remains unclear when compact adapters are sufficient, when full model updates are beneficial, and whether common capacity measures such as rank or trainable parameter count explain downstream performance.

This question is especially important for SAM-3-style segmentation models, where adaptation can affect multiple interacting components involved in visual encoding, prompting, and mask prediction. In such models, two fine-tuning methods can obtain similar validation scores while modifying different parts of the network or behaving differently under domain shift. As a result, comparing methods only by final accuracy gives an incomplete view of the adaptation process.

We organize this study around four questions about how SAM-3 adapts. First, is downstream adaptation primarily capacity-limited, such that Full Fine-Tuning should consistently outperform compact PEFT methods? Second, do nominal adapter rank and trainable parameter count reliably predict segmentation quality? Third, when PEFT methods match Full Fine-Tuning, are they simply approximating a compressed version of the Full Fine-Tuning update, or do they follow different update paths through the model? Finally, do methods with similar in-domain performance remain similar under domain shift? These questions let us separate final accuracy from the mechanisms that produce it, and they motivate our benchmark, rank analysis, update-geometry study, and LoveDA Urban/Rural transfer experiment.

In this work, we study the fine-tuning behavior of SAM-3 across five segmentation datasets: MRI-Scler, CT-Colon, CT-Vessel, LoveDA, and Apple Dense Material Segmentation~\cite{ms_lesion_challenge,medical_decathlon,loveda,apple_dms}. These datasets span medical imaging, remote sensing, and material segmentation, and include both box-prompted and text-prompted task formulations. We compare five adaptation families: LoRA, HiRA, KRAdapter, QuanTA, and Full Fine-Tuning. Our primary evaluation uses Dice similarity coefficient (DSC) and intersection-over-union (IoU), with held-out test evaluation where available.

Our results show that SAM-3 in-domain adaptation is often capacity-saturated. Compact PEFT methods frequently match or approach Full Fine-Tuning, and increasing nominal adapter rank or trainable parameter count does not consistently improve performance. We do not interpret this as evidence that PEFT is universally superior, or that SAM-3 adaptation is inherently low-rank. Rather, the results suggest that several downstream segmentation tasks admit competitive solutions through constrained adaptation paths.

To better understand this behavior, we complement the benchmark with spectral and domain-transfer analyses. The spectral analysis shows that Full Fine-Tuning updates can be compressible, but successful PEFT methods do not simply reproduce the same update structure. Instead, they can concentrate changes in different components and subspaces. The LoveDA Urban/Rural transfer study further shows that methods with similar in-domain performance can differ under domain shift, with Full Fine-Tuning giving stronger average transfer performance and LoRA showing smaller degradation in the harder transfer direction.

Our contributions are:
\begin{itemize}
\item We present a systematic SAM-3 fine-tuning benchmark across five segmentation datasets, comparing LoRA, HiRA, KRAdapter, QuanTA, and Full Fine-Tuning under both box-prompted and text-prompted protocols.
\item We show that SAM-3 in-domain adaptation is often capacity-saturated: compact PEFT methods frequently match or approach Full Fine-Tuning, while nominal adapter rank and trainable parameter count are poor standalone predictors of performance.
\item We provide spectral and domain-transfer analyses showing that competitive PEFT is not merely compressed Full Fine-Tuning, but can follow different update paths whose differences become more visible under domain shift.
\end{itemize}

\section{Related Work}

\paragraph{Promptable segmentation foundation models.}
Promptable segmentation models have shifted dense prediction from task-specific model design toward general-purpose mask prediction conditioned on user or semantic prompts. SAM introduced large-scale promptable segmentation with points, boxes, and masks as input prompts~\cite{kirillov2023segment}, and subsequent work extended the paradigm to video, stronger visual representations, and more expressive prompt interfaces~\cite{ravi2024sam2,sam3}. These models provide strong generic priors, but their zero-shot behavior can vary substantially across specialized domains, especially when the target images, object boundaries, or class definitions differ from the pretraining distribution. This makes fine-tuning strategy an important practical question for medical imaging, remote sensing, and dense material segmentation.

\paragraph{Adapting SAM-style models to specialized domains.}
A large body of work adapts SAM-style models to downstream segmentation tasks where zero-shot prompting is insufficient. In medical imaging, prior studies have explored full fine-tuning, prompt tuning, decoder tuning, and lightweight adapter insertion to improve segmentation of organs, lesions, and other domain-specific structures~\cite{medsam,samed,medical_sam_survey}. In remote sensing and industrial or material segmentation, adaptation is similarly useful because class definitions, scale, texture, and imaging conditions differ from natural-image segmentation benchmarks~\cite{remote_sam,industrial_sam}. These works show that adapting segmentation foundation models can be effective, but comparisons are often made across different datasets, prompt types, tuning protocols, and evaluation metrics. As a result, it remains difficult to determine whether improvements come from adaptation capacity, adapter placement, optimization choices, or dataset-specific effects. Our work addresses this issue by comparing multiple adaptation families under a unified SAM-3 evaluation protocol across five datasets and two prompt formulations.

\paragraph{Parameter-efficient fine-tuning.}
Parameter-efficient fine-tuning (PEFT) adapts large pretrained models while updating only a small fraction of parameters. LoRA represents task-specific changes through low-rank updates to selected weight matrices~\cite{hu2022lora}, while other structured adaptation methods introduce alternative decompositions, higher-rank parameterizations, or multiplicative/scaling-based updates to improve expressivity under a constrained parameter budget~\cite{hira,kradapter,quanta}. Although PEFT is widely used because it reduces memory and storage cost, trainable parameter count alone does not determine downstream behavior. In modular segmentation foundation models, adapters can be placed in vision encoders, prompt encoders, transformer decoders, and mask prediction modules, and different placements may lead to different adaptation paths even when final validation scores are similar. We therefore compare LoRA, HiRA, KRAdapter, QuanTA, and Full Fine-Tuning not only by final DSC and IoU, but also by rank sensitivity and component-level update behavior.

\paragraph{Low-rank structure and update geometry.}
Low-rank structure is often used to explain why compact adaptation can work well for large pretrained models. If full-model updates are concentrated in a small number of directions, then low-rank or structured adapters may be sufficient to recover much of the downstream improvement~\cite{aghajanyan2021intrinsic,lora_update_geometry}. However, strong PEFT performance does not necessarily imply that the adapter reconstructs the same solution as Full Fine-Tuning. The effective rank of the update, its localization across architectural components, and its alignment with full-model update directions can all affect downstream behavior. This distinction is especially important for SAM-3, where adaptation can occur across visual encoding, text or prompt encoding, DETR-style decoding, and mask decoding. Our study therefore complements benchmark accuracy with SVD-based update analysis and LoveDA Urban/Rural transfer experiments. This lets us test whether competitive PEFT is merely compressed Full Fine-Tuning, or whether it reaches similar in-domain performance through different update paths whose differences become visible under domain shift.

\section{Experimental Setup}
\label{sec:setup}

We evaluate the pretrained SAM-3 model and five fine-tuning families across five segmentation datasets. The pretrained model is included as a zero-shot reference to calibrate how much each dataset benefits from adaptation. The main goal of the benchmark is not to exhaustively tune every method on every dataset, but to compare representative adaptation paths under a consistent evaluation protocol. We therefore use fixed selected checkpoints for the main table, and reserve broader rank or capacity sweeps for the analysis sections.

\subsection{Datasets and task formulations}

Table~\ref{tab:datasets} summarizes the datasets and task formulations. Our benchmark includes three medical segmentation datasets, one remote-sensing dataset, and one dense material segmentation dataset. MRI-Scler, CT-Colon, and CT-Vessel are evaluated with the tracker-style SAM-3 model using box prompts; these tasks are based on public multiple-sclerosis and Medical Segmentation Decathlon resources~\cite{ms_lesion_challenge,medical_decathlon}. For tracker-style medical datasets, evaluation uses oracle box prompts derived from ground-truth masks, following common prompted-segmentation evaluation practice~\cite{kirillov2023segment,medsam}. LoveDA and Apple Dense Material Segmentation are evaluated with the text-prompted SAM-3 model, where a class name is used as the prompt and the target is the corresponding binary mask~\cite{loveda,apple_dms}.

For datasets with held-out test labels, we report test-set performance. For LoveDA, the official test split does not include labels, so we report validation performance~\cite{loveda}. For Apple Dense Material Segmentation, we use the canonical test evaluation as the main protocol~\cite{apple_dms}. This canonical evaluation enumerates valid material-class prompts and aggregates performance over the resulting prompt-mask samples. We do not use the random-one prompt protocol in the main table because it samples a single valid class prompt per image and therefore introduces additional prompt-sampling variance.

Across all datasets, the main metrics are Dice similarity coefficient (DSC) and intersection-over-union (IoU)~\cite{milletari2016vnet}. For the three medical datasets, we also report normalized surface Dice (NSD) as a secondary boundary-sensitive metric where applicable, following the Medical Segmentation Decathlon evaluation convention~\cite{medical_decathlon}. For LoveDA, we additionally use class-mean IoU in the Urban/Rural transfer analysis, following the semantic segmentation evaluator~\cite{loveda}.

\begin{table}[t]
\centering
\caption{Datasets and task formulations used in the main benchmark. LoveDA is evaluated on the validation split because test labels are unavailable. Apple DMS uses the canonical test evaluation.}
\label{tab:datasets}
\begin{tabular}{lllll}
\toprule
Dataset & Domain & SAM-3 model & Prompt & Split \\
\midrule
MRI-Scler & Medical MRI & \texttt{Sam3TrackerModel} & Box & Test \\
CT-Colon & Medical CT & \texttt{Sam3TrackerModel} & Box & Test \\
CT-Vessel & Medical CT & \texttt{Sam3TrackerModel} & Box & Test \\
LoveDA & Remote sensing & \texttt{Sam3Model} & Text & Validation \\
Apple DMS & Material segmentation & \texttt{Sam3Model} & Text & Canonical test \\
\bottomrule
\end{tabular}
\end{table}

\subsection{Adaptation methods}

We compare five fine-tuning families: Full Fine-Tuning, LoRA, HiRA, KRAdapter, and QuanTA. Full Fine-Tuning updates all model parameters and serves as the high-capacity reference. LoRA represents a standard low-rank PEFT baseline~\cite{hu2022lora}. HiRA, KRAdapter, and QuanTA provide alternative structured adaptation paths with different capacity and parameterization choices~\cite{hira,kradapter,quanta}. In addition, we report the pretrained SAM-3 model without fine-tuning as a zero-shot baseline~\cite{sam3}.

For the main benchmark, we report one checkpoint per adaptation family and dataset, selected as the best-performing configuration for that method family on the validation split. Thus, the main table compares the strongest observed LoRA, HiRA, KRAdapter, QuanTA, and Full Fine-Tuning configuration available for each dataset under the same validation-DSC checkpoint-selection rule, rather than arbitrary or manually chosen runs. This keeps the main comparison focused on adaptation families rather than a large hyperparameter grid, while the rank and capacity sweeps in Sec.~\ref{sec:rank_svd} analyze how performance changes across selected configurations. The selected main settings are summarized in Table~\ref{tab:adaptation_settings}.


\begin{table}[t]
\centering
\caption{Main adaptation settings and trainable parameter counts. Percentages are computed relative to total train-time parameters, including adapter tensors. Tracker denotes \texttt{Sam3TrackerModel}; Text denotes \texttt{Sam3Model}.}
\label{tab:adaptation_settings}
\scriptsize
\resizebox{\linewidth}{!}{
\begin{tabular}{l l r r r r}
\toprule
Method & Main setting & \multicolumn{2}{c}{Tracker} & \multicolumn{2}{c}{Text} \\
\cmidrule(lr){3-4}\cmidrule(lr){5-6}
&  & Params & \% & Params & \% \\
\midrule
Full FT & all parameters & 458.26M & 100.00 & 840.38M & 100.00 \\
LoRA & rank 8 & 2.19M & 0.48 & 4.28M & 0.51 \\
QuanTA & $d{=}4$, [8,8,4,4] & 1.32M & 0.29 & 3.14M & 0.37 \\
HiRA & rank 512 & 140.25M & 23.43 & 273.68M & 24.57 \\
KRAdapter & scaling-1 run & 8.54M & 1.83 & 15.89M & 1.86 \\
\bottomrule
\end{tabular}
}
\end{table}

For the tracker model, adapters target the attention projections in the vision backbone and mask-decoder transformer; for the text-prompted model, they target attention projections in the vision encoder, text encoder, geometry encoder, DETR encoder/decoder, and mask decoder~\cite{carion2020detr,sam3}. Heads, MLP blocks, pixel decoders, mask embedders, and scoring heads are excluded from PEFT insertion. All models are trained with AdamW, weight decay 0.1, mixed-precision bfloat16 training, a $20{:}1$ focal-to-Dice loss, 5\% linear warmup followed by cosine annealing, and checkpoint selection by validation DSC with early stopping patience 5~\cite{loshchilov2019decoupled,lin2017focal,milletari2016vnet}. PEFT runs use learning rate $10^{-4}$; Full FT uses $10^{-4}$ for heads/other modules with lower learning rates for pretrained vision/text encoders where applicable.

\subsection{Adapter placement}

The adapter target modules depend on the SAM-3 model family used by each dataset. For the tracker-style model used on MRI-Scler, CT-Colon, and CT-Vessel, adapters are inserted into attention projections in the vision backbone and the mask-decoder transformer. For the text-prompted model used on LoveDA and Apple DMS, adapters are inserted into attention projections in the vision encoder, text encoder, geometry encoder, DETR encoder/decoder, and mask decoder~\cite{carion2020detr,sam3}.

In both model families, adapters target attention projection layers rather than task heads or non-attention MLP blocks. This placement is important for the later spectral analysis, where we compare update spectra and energy localization over the same PEFT-accessible projection matrices.

\subsection{Training and evaluation protocol}

All fine-tuned models are trained with AdamW, weight decay, mixed-precision training, and a combined focal and Dice loss~\cite{loshchilov2019decoupled,lin2017focal,milletari2016vnet}. Checkpoints are selected by validation DSC. For the tracker-style datasets, box prompts are generated from ground-truth masks during training, with box jitter for augmentation. For text-prompted datasets, each training sample uses a class prompt and the corresponding binary class mask.

\paragraph{Evaluation stability.}
We repeat evaluation with three seeds to verify that the main comparisons are not driven by evaluation randomness. The canonical sweep covers all 25 checkpoint--dataset rows and is effectively deterministic: the maximum standard deviation over the three seeds is $9\times10^{-5}$ for DSC, $9\times10^{-5}$ for IoU, $1.3\times10^{-4}$ for NSD, $8\times10^{-6}$ for mIoU, and $1.9\times10^{-5}$ for mDSC. The best and worst methods by mean IoU remain unchanged for every dataset. For the DMS random-one protocol, which samples one valid material prompt per image, variance is larger but still small: on the test split, standard deviations are below $0.004$ for DSC and IoU, and below $0.015$ for mIoU. We therefore use the canonical DMS test evaluation in the main benchmark and treat random-one evaluation only as a secondary diagnostic.

\section{Main Benchmark Results}
\label{sec:main_results}

\begin{figure*}[t]
\centering
\includegraphics[width=\linewidth]{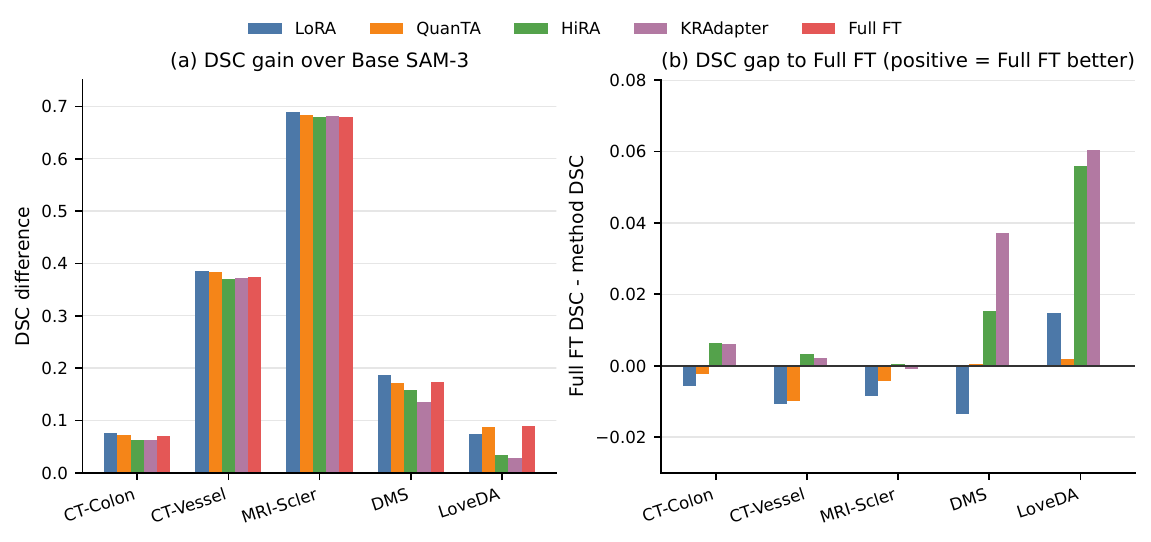}
\caption{Main benchmark summary. Left: DSC improvement over pretrained SAM-3. Right: DSC gap to Full Fine-Tuning, computed as Full FT DSC minus method DSC; positive values indicate that Full FT is better, while negative values indicate that the PEFT method is better.}
\label{fig:main_benchmark_summary}
\end{figure*}

We first compare the pretrained SAM-3 model, Full Fine-Tuning, and the four PEFT families across all five datasets. Table~\ref{tab:main_benchmark} reports DSC and IoU for the main benchmark, and Fig.~\ref{fig:main_benchmark_summary} summarizes the corresponding gains over the pretrained model and gaps to Full Fine-Tuning.

\begin{table*}[t]
\centering
\caption{Main benchmark results. Each cell reports DSC/IoU. The best DSC for each dataset is shown in bold. Base denotes pretrained SAM-3 without fine-tuning. LoveDA is evaluated on the validation split; Apple DMS uses the canonical test evaluation.}
\label{tab:main_benchmark}
\setlength{\tabcolsep}{3.69pt}
\scriptsize
\begin{tabular}{lcccccc}
\toprule
Dataset & Base & Full FT & LoRA & QuanTA & HiRA & KRAdapter \\
\midrule
CT-Colon
& 0.790/0.668
& 0.860/0.760
& \textbf{0.865/0.768}
& 0.862/0.763
& 0.853/0.751
& 0.854/0.751 \\
CT-Vessel
& 0.225/0.151
& 0.599/0.456
& \textbf{0.610/0.467}
& 0.609/0.466
& 0.596/0.455
& 0.597/0.456 \\
MRI-Scler
& 0.181/0.135
& 0.861/0.800
& \textbf{0.869/0.809}
& 0.865/0.804
& 0.860/0.800
& 0.861/0.801 \\
Apple DMS
& 0.507/0.433
& 0.680/0.589
& \textbf{0.716/0.628}
& 0.679/0.587
& 0.665/0.574
& 0.643/0.558 \\
LoveDA
& 0.541/0.430
& \textbf{0.630/0.515}
& 0.615/0.495
& 0.628/0.509
& 0.574/0.455
& 0.569/0.452 \\
\bottomrule
\end{tabular}
\end{table*}

Fine-tuning improves over the pretrained SAM-3 model on every dataset, but the size of this improvement varies substantially by domain. Base SAM-3 is already comparatively strong on CT-Colon, where it reaches 0.790 DSC, yet fine-tuning still improves the best result to 0.865 DSC. In contrast, the pretrained model performs poorly on CT-Vessel and MRI-Scler, where fine-tuning yields much larger gains. Apple DMS and LoveDA lie between these extremes: the pretrained model is not collapsed, but it remains clearly below the best fine-tuned models.

Among fine-tuned methods, compact PEFT is highly competitive with Full Fine-Tuning. LoRA achieves the best DSC on four of the five datasets: CT-Colon, CT-Vessel, MRI-Scler, and Apple DMS. QuanTA is also close to the best method in several cases, particularly on CT-Colon, CT-Vessel, and LoveDA. Full Fine-Tuning is strongest on LoveDA, but it is not uniformly dominant across the benchmark. HiRA and KRAdapter are less consistent in the selected main configurations, often trailing the strongest LoRA, QuanTA, or Full FT results.

The medical-dataset NSD results follow the same overall pattern. LoRA obtains the highest NSD on CT-Colon, CT-Vessel, and MRI-Scler, with scores of 0.781, 0.644, and 0.848, respectively. QuanTA is close on all three datasets, while Full Fine-Tuning is competitive but not always best. Thus, the DSC/IoU trends in Table~\ref{tab:main_benchmark} are not driven only by region-overlap metrics; they are also reflected in a boundary-sensitive measure.

These results support two observations. First, SAM-3 fine-tuning is not governed by a simple capacity hierarchy: updating all parameters does not consistently outperform compact adaptation. Second, the similar performance of several PEFT and Full FT configurations suggests that some datasets may admit multiple competitive adaptation paths. The CT-Colon result alone could be partly explained by the strength of the pretrained model on that dataset, but the same broad PEFT--Full FT proximity also appears on datasets where Base SAM-3 is much weaker. This motivates the rank, spectral, and transfer analyses in the following sections, where we examine whether these similar final scores arise from similar or distinct update behavior.

\section{Rank, Capacity, and Update Geometry}
\label{sec:rank_svd}

The main benchmark shows that compact PEFT methods often approach or exceed Full Fine-Tuning, even though the methods differ substantially in trainable parameters and update parameterization. In this section, we examine whether this behavior can be explained by nominal adapter capacity or by a simple low-rank approximation of Full Fine-Tuning. We find that neither explanation is sufficient: performance is often insensitive to nominal rank, and successful PEFT methods do not simply reproduce the same update structure as Full Fine-Tuning.

\subsection{Nominal rank is a weak predictor of performance}

\begin{figure*}[t]
\centering
\includegraphics[width=\linewidth]{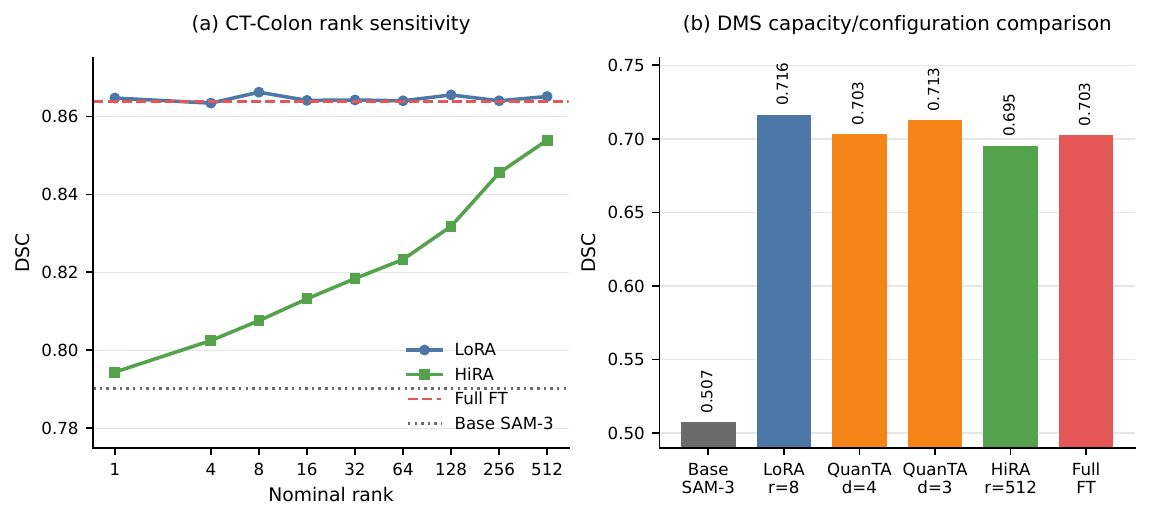}
\caption{Rank and capacity sensitivity. Left: CT-Colon rank sweep for LoRA and HiRA, with Base SAM-3 and Full Fine-Tuning shown as references. Right: Apple DMS capacity/configuration comparison. For DMS, LoRA is shown at the main rank-8 setting; the panel compares the available higher-capacity or alternative configurations for QuanTA and HiRA.}
\label{fig:rank_capacity}
\end{figure*}

Figure~\ref{fig:rank_capacity} compares adaptation capacity on CT-Colon and Apple DMS. CT-Colon provides the most complete rank sweep, with LoRA and HiRA evaluated from low to high nominal ranks. LoRA is already highly competitive at rank~1 and remains nearly flat across ranks, with rank~8 giving the strongest observed result. Increasing LoRA rank beyond this point does not produce a monotonic improvement. HiRA behaves differently: its performance improves steadily as rank increases, but even the highest-rank setting remains below the strongest LoRA and Full Fine-Tuning results on this dataset.

One possible concern is that CT-Colon may be unusually saturated because Base SAM-3 already performs well on this dataset. The DMS comparison in Fig.~\ref{fig:rank_capacity} provides a useful counterpoint. Base SAM-3 is substantially weaker on DMS, yet increasing adaptation capacity does not yield a simple ordering by method size or nominal expressivity. LoRA rank~8 remains the strongest configuration among the evaluated DMS variants, QuanTA $d=3$ and $d=4$ are close, and HiRA rank~512 does not overtake the more compact LoRA setting. Thus, the weak relationship between nominal capacity and performance is not limited to the high-base-performance CT-Colon case.

These results suggest that SAM-3 adaptation is often capacity-saturated in practice: once the method has enough flexibility to reach a good local solution, additional nominal capacity does not necessarily translate into better DSC or IoU. This does not mean that rank is irrelevant in all settings, as HiRA clearly benefits from higher rank on CT-Colon. Rather, it shows that nominal rank and trainable parameter count are poor standalone predictors of downstream segmentation performance.

\subsection{Full Fine-Tuning is compressible but not fully explained by low rank}

We next ask whether the strong performance of compact PEFT can be explained as recovering a low-rank version of the Full Fine-Tuning update. For each fine-tuned checkpoint, we analyze the weight changes over the PEFT-accessible attention projection matrices and compute their singular value spectra. We summarize each update by cumulative spectral energy, effective rank statistics, and the distribution of update energy across model components.

\begin{figure*}[t]
\centering
\includegraphics[width=\linewidth]{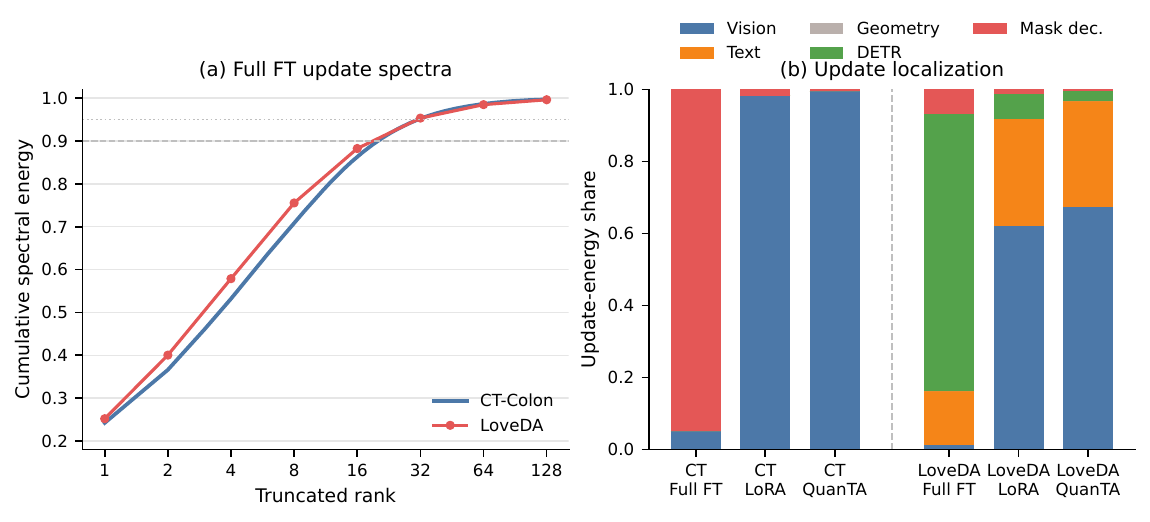}
\caption{SVD update geometry. Left: cumulative spectral energy of Full Fine-Tuning updates over selected attention projections. Right: update-energy localization across components for Full Fine-Tuning, LoRA, and QuanTA on CT-Colon and LoveDA. Full Fine-Tuning is spectrally compressible, but PEFT methods often place their update energy in different components.}
\label{fig:svd_geometry}
\end{figure*}

Figure~\ref{fig:svd_geometry}(a) shows that Full Fine-Tuning updates are spectrally compressible on both CT-Colon and LoveDA. On CT-Colon, a small number of singular directions captures a large fraction of selected-layer update energy; LoveDA shows a similar qualitative pattern. This supports the view that Full Fine-Tuning does not use all directions equally. However, compressibility alone does not imply that PEFT is simply reconstructing the Full Fine-Tuning solution. In the CT-Colon truncation experiment, low-rank reconstruction of only the selected Full FT attention-projection updates improves with rank but remains below the original Full FT checkpoint, indicating that selected low-rank reconstruction does not fully recover Full Fine-Tuning behavior.

\begin{table}[t]
\centering
\caption{Quantitative update-geometry summary for the checkpoints analyzed in Fig.~\ref{fig:svd_geometry}. Stable/effective rank, $r_{90}$, and $r_{95}$ are energy-weighted over selected PEFT-accessible projection layers. Top-component energy is the fraction of total dense update Frobenius energy assigned to the dominant component.}
\label{tab:update_geometry_summary}
\resizebox{\linewidth}{!}{
\begin{tabular}{l l r r r l r}
\toprule
Dataset & Method & Eff. rank & $r_{90}$ & $r_{95}$ & Dominant component & Top energy \\
\midrule
CT-Colon & Full FT & 4.65 & 21.97 & 33.83 & Mask decoder & 94.96\% \\
CT-Colon & LoRA $r{=}8$ & 1.49 & 3.21 & 4.59 & Vision backbone & 98.27\% \\
CT-Colon & QuanTA $d{=}4$ & 38.62 & 205.82 & 277.82 & Vision backbone & 99.54\% \\
\midrule
LoveDA & Full FT & 4.35 & 18.73 & 30.06 & DETR & 77.15\% \\
LoveDA & LoRA $r{=}8$ & 1.56 & 3.18 & 4.37 & Vision backbone & 61.99\% \\
LoveDA & QuanTA $d{=}4$ & 40.41 & 181.43 & 245.80 & Vision backbone & 67.34\% \\
\bottomrule
\end{tabular}
}
\end{table}

\subsection{PEFT and Full Fine-Tuning update different components}

The localization analysis in Fig.~\ref{fig:svd_geometry}(b) shows a stronger distinction between methods. On CT-Colon, Full Fine-Tuning concentrates most selected-layer update energy in the mask-decoder transformer, while LoRA and QuanTA concentrate most of their update energy in the vision backbone. On LoveDA, Full Fine-Tuning is dominated by DETR-related components, whereas LoRA and QuanTA place most of their update energy in the vision and text components. Geometry layers receive negligible update energy in the analyzed LoveDA checkpoints.

This mismatch is important because it shows that competitive PEFT does not need to imitate where Full Fine-Tuning changes the model. LoRA and QuanTA can reach similar or competitive performance while modifying different parts of the architecture and using different effective ranks. This also explains why alignment to the top Full FT singular subspace is not sufficient to predict downstream quality: a method can be less aligned with Full Fine-Tuning and still perform competitively if it follows a different but effective adaptation path.

Together, the rank and SVD analyses refine the interpretation of the main benchmark. Compact PEFT performs well not simply because SAM-3 updates are universally low-rank, and not because adapters recover a compressed Full Fine-Tuning solution. Instead, SAM-3 appears to admit multiple competitive adaptation paths. Full Fine-Tuning updates can be spectrally compressible, while PEFT methods may exploit different components and subspaces to reach comparable segmentation performance.

\section{Domain Transfer on LoveDA}
\label{sec:domain_transfer}

The main benchmark and update-geometry analysis show that different adaptation methods can reach similar in-domain performance through different update paths. We next ask whether these methods remain similar under domain shift. For this purpose, we use LoveDA, which provides Urban and Rural domains with distinct scene layouts and land-cover distributions~\cite{loveda}. We train source-only checkpoints on one domain and evaluate them on both Urban and Rural validation splits. This experiment is a validation-domain transfer study rather than official LoveDA test-set evaluation.

\begin{figure}[t]
\centering
\includegraphics[width=\linewidth]{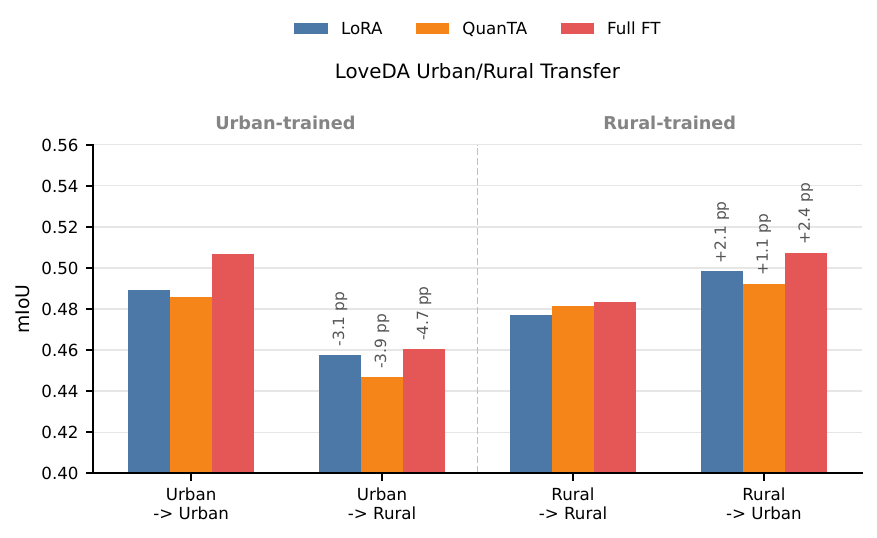}
\caption{LoveDA Urban/Rural validation-domain transfer. Models are trained on either Urban or Rural source data and evaluated on both source and target validation domains. Urban-to-Rural is the harder transfer direction: Full Fine-Tuning gives the strongest absolute cross-domain mIoU, while LoRA has the smallest drop from its in-domain score.}
\label{fig:loveda_transfer}
\end{figure}

Figure~\ref{fig:loveda_transfer} reports the Urban/Rural transfer results for LoRA, QuanTA, and Full Fine-Tuning. The two transfer directions are not symmetric. When trained on Urban and evaluated on Rural, all methods lose performance, making Urban-to-Rural the clearer stress test. In this direction, Full Fine-Tuning achieves the highest cross-domain mIoU, while LoRA shows the smallest absolute drop from its in-domain score. QuanTA is close in-domain but has the weakest Urban-to-Rural transfer among the three methods.

The Rural-to-Urban direction shows a different pattern: all methods score higher on Urban validation than on Rural validation, even when trained on Rural. We do not interpret this as evidence that cross-domain transfer improves the model. Rather, it indicates that the Urban validation split is easier under this evaluation protocol. This asymmetry is also visible in the base-model evaluation, where the pretrained model obtains higher scores on Urban than Rural. Therefore, signed domain drop should be interpreted within each transfer direction rather than compared directly across directions.

These results complement the main benchmark in two ways. First, they show that methods with similar in-domain performance can behave differently under distribution shift. Second, they clarify that there is no single dominant adaptation strategy across all criteria. Full Fine-Tuning gives the strongest average transfer performance on LoveDA, while LoRA is more stable in the harder Urban-to-Rural direction. Thus, the choice between compact PEFT and Full Fine-Tuning depends not only on in-domain DSC or IoU, but also on whether the target application prioritizes absolute transfer accuracy or robustness to domain drop.

\section{Discussion}
\label{sec:discussion}

Our results suggest that SAM-3 fine-tuning should not be viewed as a simple trade-off between parameter count and accuracy. Across the main benchmark, compact PEFT methods often match or approach Full Fine-Tuning, and in several cases outperform it. At the same time, Full Fine-Tuning remains competitive and is strongest in the LoveDA transfer analysis by average mIoU. This indicates that the best adaptation path depends on the target criterion: compact PEFT can be an effective default for in-domain performance, while Full Fine-Tuning may still be valuable when absolute transfer performance is prioritized.

The rank and SVD analyses further show that final accuracy alone can hide important differences between methods. Increasing nominal rank does not reliably improve performance, and methods with similar DSC or IoU can produce very different update structures. Full Fine-Tuning updates are spectrally compressible, but PEFT methods do not simply reconstruct the same update directions. Instead, they often concentrate update energy in different components. This supports a more nuanced interpretation of PEFT success: compact adapters can be competitive not only because they restrict the number of trainable parameters, but also because they provide alternative optimization paths through the model.

These findings lead to several practical recommendations. First, LoRA is a strong default baseline for SAM-3 adaptation: it is compact, consistently competitive, and robust across the evaluated datasets. Second, QuanTA is a useful structured alternative, especially when one wants a higher-capacity PEFT method, but our results do not show a consistent robustness advantage under domain shift. Third, Full Fine-Tuning should remain part of the comparison set whenever compute allows, since it can still provide the best performance in some settings. Finally, rank and parameter count should not be used as the only indicators of adaptation capacity; component-level update analysis provides a more informative view of how each method changes the model.

Our study also has limitations. LoveDA is evaluated on validation labels because the official test labels are unavailable, and the Urban/Rural experiment should therefore be interpreted as validation-domain transfer rather than test-set generalization. Our SVD analysis focuses on PEFT-accessible attention projections, not every parameter in the model, so it characterizes the main adapted subspaces rather than the complete full-model update. In addition, we study representative checkpoints and selected rank sweeps rather than an exhaustive hyperparameter search for every method on every dataset. Despite these constraints, the consistency of the benchmark, seed-stability checks, spectral analysis, and domain-transfer study provides a coherent picture: SAM-3 admits multiple competitive fine-tuning paths, and their differences are better understood through both performance and update geometry.

\section{Conclusion}
\label{sec:conclusion}

We presented a systematic study of SAM-3 fine-tuning across five segmentation datasets and five adaptation families. Our results show that compact PEFT methods often match or approach Full Fine-Tuning in DSC and IoU, even when the pretrained model itself varies substantially in zero-shot performance across datasets. This indicates that strong SAM-3 adaptation does not always require updating the full model.

Beyond final accuracy, our analysis shows that this behavior is not explained by nominal rank or trainable parameter count alone. Rank sweeps and DMS capacity comparisons reveal that larger nominal capacity does not consistently improve performance. SVD analysis further shows that Full Fine-Tuning updates can be spectrally compressible, but successful PEFT methods often concentrate their updates in different components and subspaces rather than simply approximating the Full Fine-Tuning solution. Finally, the LoveDA Urban/Rural transfer study shows that methods with similar in-domain performance can differ under domain shift.

Together, these findings suggest that SAM-3 admits multiple competitive fine-tuning paths. Compact adapters such as LoRA provide strong default baselines, structured PEFT methods such as QuanTA can be competitive alternatives, and Full Fine-Tuning remains important in settings where absolute transfer performance is the priority. More broadly, our results suggest that future work on segmentation foundation model adaptation should evaluate not only final accuracy, but also update geometry, capacity sensitivity, and robustness under domain shift.

\section*{Acknowledgments}

This work was supported by the NVIDIA Academic Grant Program.
We gratefully acknowledge NVIDIA Corporation for providing
32,000 NVIDIA A100 GPU-hours, which supported the experiments
presented in this work.

%
%
\bibliographystyle{splncs04}
\bibliography{main}

@inproceedings{kirillov2023segment,
  title     = {Segment Anything},
  author    = {Kirillov, Alexander and Mintun, Eric and Ravi, Nikhila and Mao, Hanzi and Rolland, Chloe and Gustafson, Laura and Xiao, Tete and Whitehead, Spencer and Berg, Alexander C. and Lo, Wan-Yen and Doll{\'a}r, Piotr and Girshick, Ross},
  booktitle = {Proceedings of the IEEE/CVF International Conference on Computer Vision (ICCV)},
  pages     = {4015--4026},
  year      = {2023},
  doi       = {10.48550/arXiv.2304.02643},
  url       = {https://arxiv.org/abs/2304.02643},
  note      = {Accessed 30 Sep 2026}
}

@article{ravi2024sam2,
  title   = {{SAM} 2: Segment Anything in Images and Videos},
  author  = {Ravi, Nikhila and Gabeur, Valentin and Hu, Yuan-Ting and Hu, Ronghang and Ryali, Chaitanya and Ma, Tengyu and Khedr, Haitham and R{\"a}dle, Roman and Rolland, Chloe and Gustafson, Laura and Mintun, Eric and Pan, Junting and Alwala, Kalyan Vasudev and Carion, Nicolas and Wu, Chao-Yuan and Girshick, Ross and Doll{\'a}r, Piotr and Feichtenhofer, Christoph},
  journal = {arXiv preprint arXiv:2408.00714},
  year    = {2024},
  doi     = {10.48550/arXiv.2408.00714},
  url     = {https://arxiv.org/abs/2408.00714},
  note    = {Accessed 30 Sep 2026}
}

@article{sam3,
  title   = {{SAM} 3: Segment Anything with Concepts},
  author  = {Carion, Nicolas and Gustafson, Laura and Hu, Yuan-Ting and Debnath, Shoubhik and Hu, Ronghang and Suris, Didac and Ryali, Chaitanya and Alwala, Kalyan Vasudev and Khedr, Haitham and Huang, Andrew and Lei, Jie and Ma, Tengyu and Guo, Baishan and Kalla, Arpit and Marks, Markus and Greer, Joseph and Wang, Meng and Sun, Peize and R{\"a}dle, Roman and Afouras, Triantafyllos and Mavroudi, Effrosyni and Xu, Katherine and Wu, Tsung-Han and Zhou, Yu and Momeni, Liliane and Hazra, Rishi and Ding, Shuangrui and Vaze, Sagar and Porcher, Francois and Li, Feng and Li, Siyuan and Kamath, Aishwarya and Cheng, Ho Kei and Doll{\'a}r, Piotr and Ravi, Nikhila and Saenko, Kate and Zhang, Pengchuan and Feichtenhofer, Christoph},
  journal = {arXiv preprint arXiv:2511.16719},
  year    = {2025},
  doi     = {10.48550/arXiv.2511.16719},
  url     = {https://arxiv.org/abs/2511.16719},
  note    = {Accessed 30 Sep 2026}
}

@article{medsam,
  title   = {Segment Anything in Medical Images},
  author  = {Ma, Jun and He, Yuting and Li, Feifei and Han, Lin and You, Chenyu and Wang, Bo},
  journal = {Nature Communications},
  volume  = {15},
  number  = {1},
  pages   = {654},
  year    = {2024},
  doi     = {10.1038/s41467-024-44824-z},
  url     = {https://doi.org/10.1038/s41467-024-44824-z},
  note    = {Accessed 30 Sep 2026}
}

@article{samed,
  title   = {Customized Segment Anything Model for Medical Image Segmentation},
  author  = {Zhang, Kaidong and Liu, Dong},
  journal = {arXiv preprint arXiv:2304.13785},
  year    = {2023},
  doi     = {10.48550/arXiv.2304.13785},
  url     = {https://arxiv.org/abs/2304.13785},
  note    = {Accessed 30 Sep 2026}
}

@article{medical_sam_survey,
  title   = {Segment Anything Model for Medical Images?},
  author  = {Huang, Yuhao and Yang, Xin and Liu, Lian and Zhou, Han and Chang, Ao and Zhou, Xinrui and Chen, Rusi and Yu, Junxuan and Chen, Jiongquan and Chen, Chaoyu and Liu, Sijing and Chi, Haozhe and Hu, Xindi and Yue, Kejuan and Li, Lei and Grau, Vicente and Fan, Deng-Ping and Dong, Fajin and Ni, Dong},
  journal = {arXiv preprint arXiv:2304.14660},
  year    = {2023},
  doi     = {10.48550/arXiv.2304.14660},
  url     = {https://arxiv.org/abs/2304.14660},
  note    = {Accessed 30 Sep 2026}
}

@article{remote_sam,
  title   = {{RSPrompter}: Learning to Prompt for Remote Sensing Instance Segmentation Based on Visual Foundation Model},
  author  = {Chen, Keyan and Liu, Chenyang and Chen, Hao and Zhang, Haotian and Li, Wenyuan and Zou, Zhengxia and Shi, Zhenwei},
  journal = {IEEE Transactions on Geoscience and Remote Sensing},
  volume  = {62},
  pages   = {1--17},
  year    = {2024},
  doi     = {10.1109/TGRS.2024.3356074},
  url     = {https://arxiv.org/abs/2306.16269},
  note    = {Accessed 30 Sep 2026}
}

@article{industrial_sam,
  title   = {{SAM} Era: Can It Segment Any Industrial Surface Defects?},
  author  = {Song, Kechen and Cui, Wenqi and Yu, Han and Li, Xingjie and Yan, Yunhui},
  journal = {Computers, Materials \& Continua},
  volume  = {78},
  number  = {3},
  pages   = {3953--3969},
  year    = {2024},
  doi     = {10.32604/cmc.2024.048451},
  url     = {https://doi.org/10.32604/cmc.2024.048451},
  note    = {Accessed 30 Sep 2026}
}

@inproceedings{hu2022lora,
  title     = {{LoRA}: Low-Rank Adaptation of Large Language Models},
  author    = {Hu, Edward J. and Shen, Yelong and Wallis, Phillip and Allen-Zhu, Zeyuan and Li, Yuanzhi and Wang, Shean and Wang, Lu and Chen, Weizhu},
  booktitle = {International Conference on Learning Representations},
  year      = {2022},
  url       = {https://openreview.net/forum?id=nZeVKeeFYf9},
  note      = {Accessed 30 Sep 2026}
}

@inproceedings{hira,
  title     = {Hi{RA}: Parameter-Efficient Hadamard High-Rank Adaptation for Large Language Models},
  author    = {Huang, Qiushi and Ko, Tom and Zhuang, Zhan and Tang, Lilian and Zhang, Yu},
  booktitle = {The Thirteenth International Conference on Learning Representations},
  year      = {2025},
  url       = {https://openreview.net/forum?id=TwJrTz9cRS},
  note      = {Accessed 30 Sep 2026}
}

@inproceedings{kradapter,
  title     = {Towards Higher Effective Rank in Parameter-Efficient Fine-Tuning using Khatri--Rao Product},
  author    = {Albert, Paul and Zhang, Frederic Z. and Saratchandran, Hemanth and van den Hengel, Anton and Abbasnejad, Ehsan},
  booktitle = {Proceedings of the IEEE/CVF International Conference on Computer Vision (ICCV)},
  year      = {2025},
  doi       = {10.48550/arXiv.2508.00230},
  url       = {https://arxiv.org/abs/2508.00230},
  note      = {Accessed 30 Sep 2026}
}

@inproceedings{quanta,
  title     = {{QuanTA}: Efficient High-Rank Fine-Tuning of {LLM}s with Quantum-Informed Tensor Adaptation},
  author    = {Chen, Zhuo and Dangovski, Rumen and Loh, Charlotte and Dugan, Owen and Luo, Di and Solja{\v c}i{\'c}, Marin},
  booktitle = {Advances in Neural Information Processing Systems},
  year      = {2024},
  doi       = {10.48550/arXiv.2406.00132},
  url       = {https://arxiv.org/abs/2406.00132},
  note      = {Accessed 30 Sep 2026}
}

@inproceedings{aghajanyan2021intrinsic,
  title     = {Intrinsic Dimensionality Explains the Effectiveness of Language Model Fine-Tuning},
  author    = {Aghajanyan, Armen and Zettlemoyer, Luke and Gupta, Sonal},
  booktitle = {Proceedings of the 59th Annual Meeting of the Association for Computational Linguistics and the 11th International Joint Conference on Natural Language Processing},
  pages     = {7319--7328},
  year      = {2021},
  doi       = {10.18653/v1/2021.acl-long.568},
  url       = {https://aclanthology.org/2021.acl-long.568},
  note      = {Accessed 30 Sep 2026}
}

@article{lora_update_geometry,
  title   = {{DoRA}: Weight-Decomposed Low-Rank Adaptation},
  author  = {Liu, Shih-Yang and Wang, Chien-Yi and Yin, Hongxu and Molchanov, Pavlo and Wang, Yu-Chiang Frank and Cheng, Kwang-Ting and Chen, Min-Hung},
  journal = {arXiv preprint arXiv:2402.09353},
  year    = {2024},
  doi     = {10.48550/arXiv.2402.09353},
  url     = {https://arxiv.org/abs/2402.09353},
  note    = {Accessed 30 Sep 2026}
}

@article{medical_decathlon,
  title   = {The Medical Segmentation Decathlon},
  author  = {Antonelli, Michela and Reinke, Annika and Bakas, Spyridon and Farahani, Keyvan and Kopp-Schneider, Annette and Landman, Bennett A. and Litjens, Geert and Menze, Bjoern and Ronneberger, Olaf and Summers, Ronald M. and van Ginneken, Bram and others},
  journal = {Nature Communications},
  volume  = {13},
  number  = {1},
  pages   = {4128},
  year    = {2022},
  doi     = {10.1038/s41467-022-30695-9},
  url     = {https://doi.org/10.1038/s41467-022-30695-9},
  note    = {Accessed 30 Sep 2026}
}

@article{ms_lesion_challenge,
  title   = {Longitudinal Multiple Sclerosis Lesion Segmentation: Resource and Challenge},
  author  = {Carass, Aaron and Roy, Snehashis and Jog, Amod and Cuzzocreo, Jennifer L. and Magrath, Elizabeth and Gherman, Adrian and Button, Jennifer and Nguyen, Jessie and Prados, Ferran and Sudre, Carole H. and others},
  journal = {NeuroImage},
  volume  = {148},
  pages   = {77--102},
  year    = {2017},
  doi     = {10.1016/j.neuroimage.2016.12.064},
  url     = {https://doi.org/10.1016/j.neuroimage.2016.12.064},
  note    = {Accessed 30 Sep 2026}
}

@inproceedings{loveda,
  title     = {{LoveDA}: A Remote Sensing Land-Cover Dataset for Domain Adaptive Semantic Segmentation},
  author    = {Wang, Junjue and Zheng, Zhuo and Ma, Ailong and Lu, Xiaoyan and Zhong, Yanfei},
  booktitle = {Proceedings of the Neural Information Processing Systems Track on Datasets and Benchmarks},
  year      = {2021},
  doi       = {10.48550/arXiv.2110.08733},
  url       = {https://arxiv.org/abs/2110.08733},
  note      = {Accessed 30 Sep 2026}
}

@inproceedings{apple_dms,
  title     = {A Dense Material Segmentation Dataset for Indoor and Outdoor Scene Parsing},
  author    = {Upchurch, Paul and Niu, Ransen},
  booktitle = {European Conference on Computer Vision (ECCV)},
  year      = {2022},
  doi       = {10.48550/arXiv.2207.10614},
  url       = {https://machinelearning.apple.com/research/dense-material},
  note      = {Accessed 30 Sep 2026}
}

@inproceedings{carion2020detr,
  title     = {End-to-End Object Detection with Transformers},
  author    = {Carion, Nicolas and Massa, Francisco and Synnaeve, Gabriel and Usunier, Nicolas and Kirillov, Alexander and Zagoruyko, Sergey},
  booktitle = {European Conference on Computer Vision},
  pages     = {213--229},
  year      = {2020},
  organization = {Springer}
}

@inproceedings{loshchilov2019decoupled,
  title     = {Decoupled Weight Decay Regularization},
  author    = {Loshchilov, Ilya and Hutter, Frank},
  booktitle = {International Conference on Learning Representations},
  year      = {2019},
  url       = {https://openreview.net/forum?id=Bkg6RiCqY7},
  note      = {Accessed 30 Sep 2026}
}

@inproceedings{lin2017focal,
  title     = {Focal Loss for Dense Object Detection},
  author    = {Lin, Tsung-Yi and Goyal, Priya and Girshick, Ross and He, Kaiming and Doll{\'a}r, Piotr},
  booktitle = {Proceedings of the IEEE International Conference on Computer Vision},
  pages     = {2980--2988},
  year      = {2017}
}

@inproceedings{milletari2016vnet,
  title     = {{V-Net}: Fully Convolutional Neural Networks for Volumetric Medical Image Segmentation},
  author    = {Milletari, Fausto and Navab, Nassir and Ahmadi, Seyed-Ahmad},
  booktitle = {2016 Fourth International Conference on 3D Vision (3DV)},
  pages     = {565--571},
  year      = {2016},
  organization = {IEEE},
  doi       = {10.1109/3DV.2016.79}
}
\end{document}